\documentclass{article}

\usepackage{microtype}
\usepackage{graphicx}
\usepackage{booktabs} 

\usepackage{hyperref}

\usepackage[accepted]{icml2025}

\usepackage{amsmath}
\usepackage{amssymb}
\usepackage{mathtools}
\usepackage{amsthm}

\usepackage[capitalize,noabbrev]{cleveref}

\theoremstyle{plain}

\theoremstyle{definition}

\theoremstyle{remark}

\newcommand{\speedx}{{\ooalign{$\phantom{0}$\cr\hidewidth$\scriptstyle\times$\cr}}}

\usepackage{algorithm}
\usepackage{algorithmic}
\usepackage{colortbl}
\usepackage{caption}
\usepackage{multirow}
\usepackage{lipsum}
\usepackage{here}
\usepackage{amsmath}
\usepackage{amssymb}
\usepackage{bm}
\usepackage{subcaption}
\usepackage{enumitem}
\usepackage[capitalize]{cleveref}
\usepackage{sidecap}
\usepackage{wrapfig}
\usepackage{float}
\usepackage{siunitx}

\setlist[itemize]{leftmargin=*,nosep}
\setlist[enumerate]{leftmargin=*,nosep}
\setlist[description]{leftmargin=1em,nosep}
\newcolumntype{L}{>{\hspace*{-\tabcolsep}}l}
\newcolumntype{R}{r<{\hspace*{-\tabcolsep}}}

\DeclareMathOperator{\sign}{sign}

\usepackage[textsize=tiny]{todonotes}

\icmltitlerunning{Lion Cub: Minimizing Communication Overhead in Distributed Lion}

\begin{document}

\twocolumn[
\icmltitle{Lion Cub: Minimizing Communication Overhead in Distributed Lion}




\begin{icmlauthorlist}
\icmlauthor{Satoki Ishikawa}{st}
\icmlauthor{Tal Ben-Nun}{lv}
\icmlauthor{Brian Van Essen}{lv}
\icmlauthor{Rio Yokota}{st}
\icmlauthor{Nikoli Dryden}{lv}
\end{icmlauthorlist}

\icmlaffiliation{st}{Institute of Science Tokyo}
\icmlaffiliation{lv}{Lawrence Livermore National Laboratory}

\icmlcorrespondingauthor{Satoki Ishikawa}{ishikawa@rio.scrc.iir.isct.ac.jp}

\icmlkeywords{Machine Learning, ICML}

\vskip 0.3in
]



\printAffiliationsAndNotice{\icmlEqualContribution} 

\begin{abstract}
Communication overhead is a key challenge in distributed deep learning, especially on lower-bandwidth interconnects, and given current hardware trends, communication is likely to become a major bottleneck.
While gradient compression techniques have been explored for SGD and Adam, the Lion optimizer has the distinct advantage that its update vectors are the output of a sign operation, enabling straightforward quantization.
However, simply compressing updates for communication and using techniques like majority voting fails to lead to end-to-end speedups due to inefficient communication algorithms and reduced convergence.
We analyze three factors critical to distributed learning with Lion: optimizing communication methods, identifying effective quantization methods, and assessing the necessity of momentum synchronization.
Our findings show that quantization techniques adapted to Lion and selective momentum synchronization can reduce communication costs while maintaining convergence.
We combine these into Lion Cub, which enables speedups in training compared to Lion especially with low-bandwidth environments.
This highlights Lion's potential as a communication-efficient solution for distributed training.
\end{abstract}

\section{Introduction}

Recent progress in deep learning has largely been driven by improvements in computational power~\citep{kaplan2020scaling,hoffmann2022training}.
Training large-scale models on distributed systems using hundreds or thousands of GPUs has become standard practice.
This shift has increased the demand for scalable and communication-efficient distributed training algorithms capable of handling large-scale systems.
Thus, a key challenge to train large-scale models is minimizing communication overhead, which frequently limits the scalability of distributed training.

To date, communication has typically been optimized via efficient allreduce algorithms and frameworks~\citep[e.g.,][]{thakur2005optimization,nccl,cai2021synthesizing}, communication/computation overlap, or through improved network hardware and increased numbers of NICs per compute node.
However, current trends show computational throughput growing much faster than communication bandwidth.
For example, the InfiniBand roadmap indicates 2\speedx{} bandwidth growth every four years, whereas Nvidia has delivered 2--3\speedx{} flop increases every two years~\citep{ibroadmap,nvroadmap}.
While technologies such as NVLink allow for higher-bandwidth cliques, these do not scale to entire clusters.
Further, the growth in model size far outpaces the growth in GPU memory capacity, requiring smaller per-GPU batch sizes.
As communication volume is proportional to the model size and computation to the batch size, this leads to further imbalance. %
The imbalance is even larger in loosely coupled grid systems, where workers may be separated by wide‑area links, and in edge or collaborative learning settings that rely on wireless channels with modest throughput and variable latency~\citep{zhang2020network,diskin2021distributed,ryabinin2021moshpit,hilmkil2021scaling,borzunov2022petals}.
Consequently, the long-term trend is toward communication costs dominating computation.

\begin{figure*}[ht]%
    \centering%
    \includegraphics[width=\linewidth]{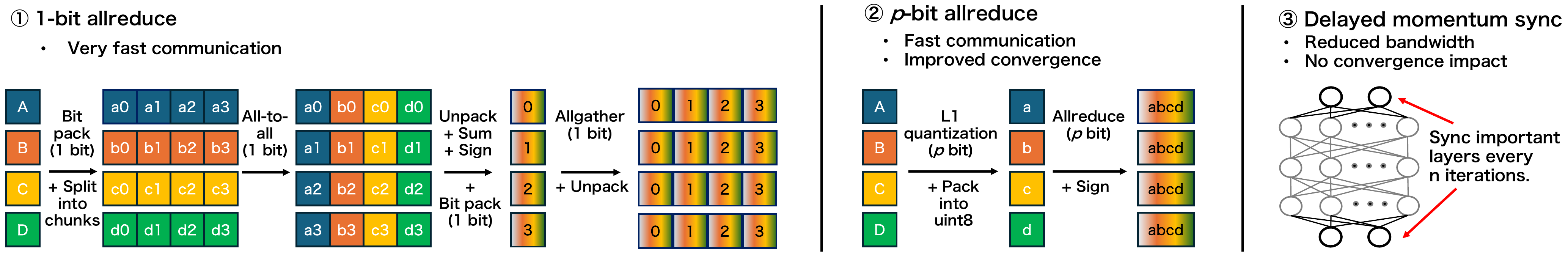}%
    \caption{\textbf{Overview of our communication algorithm.}
    We propose two synchronization methods for Lion’s update vector: 1-bit allreduce and $p$-bit allreduce.
1-bit allreduce provides fast communication but may compromise convergence. Additionally, the frequent packing and unpacking of 1-bit data can introduce overhead. In contrast, $p$-bit allreduce requires more communication time than 1-bit allreduce but typically converges more quickly. As a result, $p$-bit allreduce can sometimes achieve faster convergence of the training loss relative to training time.
Synchronizing momentum can also help maintain convergence stability.}\label{fig:overview-figure1}%
\end{figure*}

Many algorithmic approaches have been developed to reduce communication volume, typically via gradient compression~\citep{tang2020communication} in the context of SGD\@.
These include sign-based algorithms~\citep[e.g.,][]{seide20141,bernstein2018signsgd,bernstein2018signsgdmaj} that communicate only gradient sign; sparsification~\citep[e.g.,][]{strom2015scalable,dryden2016communication,lin2018deep,renggli2019sparcml}; and randomized compression operators~\citep[e.g.,][]{alistarh2017qsgd,wen2017terngrad,wang2018atomo}.
Similar ideas have been applied to more complex optimizers, such as Adam or LAMB~\citep{tang21a,li20221,lu2023maximizing}.
However, these methods often slow down convergence due to information loss during compression, although recent papers have mitigated this effect.

In contrast to these approaches, we focus on the Lion optimizer~\citep{chen2024symbolic}, which directly applies the $\sign$ operator to its updates while maintaining performance comparable to AdamW on a wide range of tasks.
Hence, Lion is a promising candidate for gradient compression to reduce communication overhead.
Recently, \citet{liu2024communication} analyzed distributed Lion and theoretically demonstrated that it converges with a ``majority vote'' update rule, and validated this experimentally.
However, their communication implementation is based on parameter servers, which do not scale, and they do not demonstrate that distributed Lion can achieve a real-time speedup for training.

\Cref{fig:overview-figure1} is an overview of our key contributions.
We first study the communication primitives required for distributed Lion, and implement a 1-bit allreduce similar to that of \citet{tang21a}, which significantly reduces bandwidth requirements.
We then analyze the distribution of gradient updates across layers in Lion and the impact of the 1-bit allreduce, and identify key limitations of the 1-bit representation, including occasional loss spikes during training.
To remedy this, we develop a novel L1 quantization method for Lion, which can be efficiently implemented with a $p$-bit allreduce while significantly improving convergence.
We also study the impact of distributed and compressed communication on momentum in Lion, and find that local estimates of momentum are sufficient except for certain layers, which may require occasional momentum synchronization to ensure convergence.
Finally, we experimentally evaluate our implementation and show that distributed Lion can deliver improved wall-time training on ResNet and transformer models without compromising model quality.

We summarize our contributions as follows:
\begin{itemize}[itemsep=0.1mm,parsep=0pt]
\item We propose optimized communication primitives to implement both 1-bit and $p$-bit allreduces specialized for distributed Lion, which outperform parameter servers.
\item We analyze the distribution of gradient updates in Lion, and show that standard quantization methods are ineffective due to outliers. We propose an L1 quantization method which offers superior performance while still enabling efficient communication.
\item We analyze the necessity of momentum synchronization in Lion and find that it can often be eliminated or performed only periodically on key layers.
\item We experimentally evaluate our distributed Lion implementation and improve training wall-time by up to 5.1\speedx{} without impacting quality for large models.
\end{itemize}

\section{Background and Related Work}
\paragraph{Lion}

The Lion optimizer~\citep{chen2024symbolic} is a recent memory-efficient optimizer with performance comparable to AdamW.
It has the following update rule:
\begin{align*}
m_{t+1} & =\beta_2 m_t-\left(1-\beta_2\right) \nabla f\left(\theta_t\right) \\
\theta_{t+1} & =\theta_t+ \eta_t \left(\sign\left(\beta_1 m_t-\left(1-\beta_1\right) \nabla f\left(\theta_t\right)\right)\right),
\end{align*}
where $\eta_t$ is the learning rate, $\beta_1$ and $\beta_2$ are hyperparameters
, and $m_t$ the momentum.
When $\beta_1 = \beta_2$, Lion matches sign descent.
However, when $\beta_2 > \beta_1$, the importance of the current gradient $\nabla f_t$ is increased compared to sign descent with momentum~\citep{liu2024communication}.
Recently, \citet{liu2024communication} proposed Distributed Lion, which combines Lion with majority voting, and demonstrated its convergence both theoretically and experimentally.

\paragraph{1-bit Communication}
There has also been research aimed at reducing communication costs by applying 1-bit communication to optimizers other than SGD\@.
1-bit Adam and 1-bit LAMB~\citep{tang21a, li20221} propose methods for compressing update (momentum) vectors for Adam and LAMB\@.
As these optimizers do not use a $\sign$ or similar operation, directly compressing the updates results in accuracy degradation, and incorporate methods such as error feedback~\citep{seide20141} to compensate.
Other systems, such as QSGD~\citep{alistarh2017qsgd}, have generalized 1-bit communication to arbitrary bitwidths.

\paragraph{Quantization}
When quantizing to $n$ bits, it is essential to ensure the full range of the representation is used.
This is typically achieved by normalizing and rescaling the input data by the absolute maximum of the input elements, e.g.:
$Q_{\infty}(x) = \operatorname{sround} \left( \frac{2^{n-1} -1 }{\| x \|_{\infty}} x \right)$ 
where $\| \cdot \|_{\infty}$ is the maximum norm and $\operatorname{sround}$ is a stochastic rounding operator~\citep{alistarh2017qsgd}.
This simple quantization method performs well with a uniform distribution, but frequently fails when applied to data following other distributions~\citep{dettmers2022bit, dettmers2023qlora}.
Empirical evidence shows that gradients in deep learning frequently follow a Laplace or lognormal distribution, leading to methods specifically tailored to these distributions~\citep[e.g.,][]{Yu2020,chmiel2020neural} but which are computationally intensive.

\section{Communication Design}



\begin{table}[t]
  \centering
  \scriptsize
  \renewcommand{\arraystretch}{1.2}
  \setlength{\tabcolsep}{3pt}
  \begin{tabular}{LrR}
    \toprule
    Implementation & Latency cost & Bandwidth cost \\
    \midrule
    Parameter server (na\"{\i}ve)   & $2(P-1)\,\alpha$                     & $2P\,N\,b\,\beta$ \\
    Parameter server (efficient)    & $2\log_2(P)\,\alpha$                 & $3\frac{P-1}{P}N\,b\,\beta$ \\
    Direct allreduce                & $2\log_2(P)\,\alpha$                 & $2\frac{P-1}{P}N(\log_2(P)+1)\,\beta$ \\
    1-bit compressed allreduce      & $(P-1+\log_2(P))\,\alpha$            & $\bigl(1+\frac{P-1}{P}\bigr)N\,\beta$ \\
    \bottomrule
  \end{tabular}
  \caption[Comparison of communication bandwidth costs.]{%
    Comparison of communication bandwidth costs for implementing majority voting with
    $P$ workers, $N$ parameters, $b$ bits per standard word, latency $\alpha$, and inverse bandwidth $\beta$ (s/bit).
  }
  \label{tab:majvote-cost}
\end{table}

\begin{figure}[t]
    \centering%
    \includegraphics[width=\linewidth]{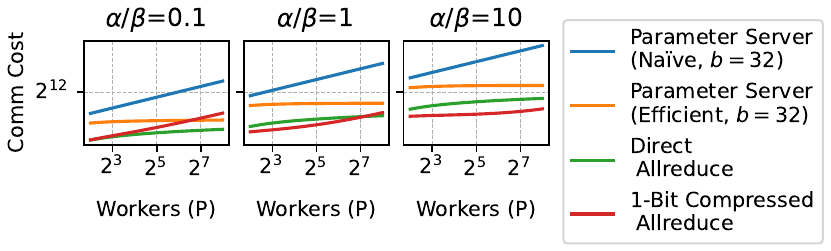}%
    \caption{\textbf{Communication costs with our performance model.}
      We illustrate the communication costs from~\cref{tab:majvote-cost} for different cases of relative latency and bandwidth.
      }
      \label{fig:band-width}%
\end{figure}

The communication for majority voting can be implemented in several ways.
We compare three implementations: parameter servers, a direct allreduce, and a variant of a 1-bit compressed allreduce.
We use the $\alpha$-$\beta$ model, where $\alpha$ is the latency in seconds and $\beta$ the inverse bandwidth in seconds per bit.
Let there be $P$ workers and $N$ parameters, and assume there are $b$ bits in a word.
See, e.g., \citet{thakur2005optimization,chan2007collective} for derivations of the communication costs we use.
We focus on bandwidth in the text and summarize both latency and bandwidth terms in \cref{tab:majvote-cost}.
For simplicity, we ignore computational overheads.

In a parameter server model, as assumed by distributed Lion~\citep{liu2024communication}, each worker sends its update to a centralized parameter server via a gather operation.
The server then aggregates the updates and broadcasts the final update to the workers.
A na\"{\i}ve implementation has each worker send to the server for cost $PNb\beta$.
A similar implementation can be used for the broadcast, again requiring $PNb\beta$, for a total cost of $2PNb\beta$.
A more efficient implementation of the gather and broadcast could instead be used, with cost $3\frac{P-1}{P}Nb\beta$.
However, allreduce-based approaches are more efficient, and further, do not risk communication being bottlenecked at a single server.

In a direct allreduce implementation, we first convert signs to 0/1 (forcing 0 to be one or the other; see the sequel for a discussion) and then performing a standard allreduce with summation to obtain the majority vote.
As the sum is between $0$ and $P$, we require $\log_2(P) + 1$ bits per parameter.
This allreduce requires time $2 \frac{P-1}{P} N (\log_2(P) + 1) \beta$, a savings of $\frac{b}{\log_2(P) + 1}$ by using a more concise representation.
In practice, we typically use 8-bit words and pack entries.

Lastly, we can adapt the 1-bit compressed allreduce of \citet{tang21a} to support majority voting.
The update vector is first represented to 1 bit (again forcing 0 to be either -1 or 1).
This is then split into $P$ equal chunks, and the $i$th chunk is sent to the $i$th worker via an all-to-all communication.
Each worker locally sums the contributions, takes the sign, and recompresses the result to 1 bit.
Finally, an allgather is performed to disseminate the complete update to every worker.
(See \cref{fig:overview-figure1} for an illustration.)
We assume the all-to-all is implemented as a pairwise exchange with cost $N \beta$, and the allgather has cost $\frac{P-1}{P} N \beta$, yielding the lowest bandwidth cost, although care must be taken to efficiently implement the intermediate packing and unpacking.

\Cref{fig:band-width} provides an illustrative comparison of these algorithms for different scales and latency/bandwidth ratios.
The na\"{\i}ve parameter server scales poorly in all situations.
The direct allreduce offers the best performance in high-latency settings and at large scale, where the cost of the all-to-all makes the 1-bit compressed allreduce less efficient.
Conversely, in bandwidth-bound regimes, the lower data transfer requirements of the 1-bit allreduce are more advantageous.

\section{Quantization Design}


\subsection{The distribution of update vectors}

\begin{figure}[t]%
    \centering%
    \includegraphics[width=\linewidth]{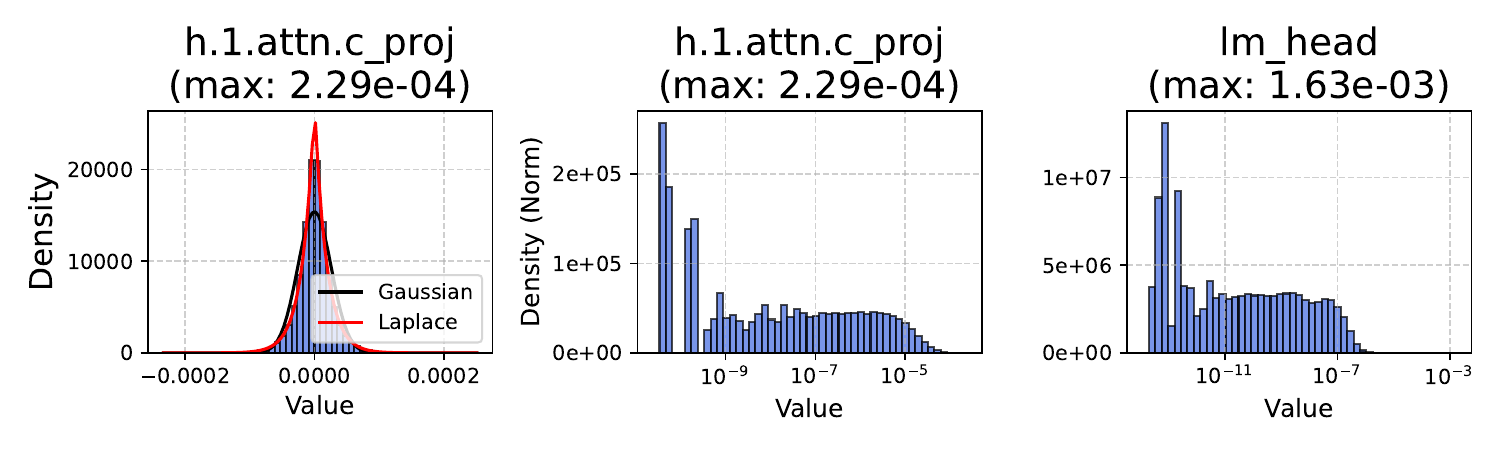}%
    \caption{\textbf{The distribution of the update vector is not uniform.}
      The distribution represents the values of the update vector before quantization ($\sign$).
      Left: The values do not follow a uniform distribution; rather, they approximate a Laplace distribution (not a Gaussian distribution).
      Right: The norm of the update vector is heavy-tailed, with the maximum significantly exceeding the mean.
      }\label{fig:update_distribution}%
\end{figure}

The commonly used quantization function $Q_{\infty}(x)$ is information theoretically optimal when $x$ follows a uniform distribution.
However, it can introduce significant error when the distribution of $x$ deviates too far from uniform.
In particular, it is sensitive to outliers: when the maximum value is excessively large, most values are quantized to zero.

Figure~\ref{fig:update_distribution} illustrates the distribution of the update vector. 
This distribution is not uniform and is more closely aligned with a Laplace distribution than a Gaussian distribution. Consequently, when standard stochastic quantization is applied, the presence of outlier values results in most values being quantized to zero.
In fact, for the last layer (lm\_head), the values of the update vector mostly range from $1 \times 10^{-12}$ to 1 $\times 10^{-7}$. However, due to the maximum value of $1.63 \times 10^{-3}$, quantizing with 5 bits leads to any value below $1 \times 10^{-4}$ being represented as zero, thus most values are quantized into 0. 
This indicates the limitation of applying standard stochastic quantization to Lion.

\subsection{L1 Quantization algorithm}

\begin{table}[]   
  \scriptsize
  \centering
  \renewcommand{\arraystretch}{1.2}
  \setlength{\tabcolsep}{8pt}
  \caption{\textbf{Validation loss on the Fineweb dataset with different quantization function.}}
  \label{tab:optimizer_loss}
  \begin{tabular}{lcccccc}
    \toprule
    \textbf{Optim} 
      & \textbf{32\,bit} 
      & \textbf{1\,bit} 
      & \textbf{8\,bit~($L_0$)} 
      & \textbf{8\,bit~($L_\infty$)} 
      & \textbf{8\,bit~($L_1$)} \\
    \midrule
    \textbf{Loss} 
      & 2.94 
      & 3.22 
      & 3.07 
      & 4.64 
      & 2.97 \\
    \bottomrule
  \end{tabular}
\end{table}

We consider a new type of quantization function that normalizes by the $L_p$ norm instead of the $L_{\infty}$ norm, making it robust against outliers:
\begin{equation}
    Q_{p}(x)_{i}
=
\operatorname{clamp}\!\Biggl(
  \operatorname{round}\!\Bigl(
    \frac{2^{n-1}-1}
         {2 M_{p}(x)}
    \,x_{i}
  \Bigr),\,2^{n-1}-1 \Biggr),
\end{equation}
where $M_{p}(x)
= \Bigl(\tfrac{1}{d}\sum_{j=1}^{d}|x_{j}|^{p}\Bigr)^{\!1/p}$ and $\operatorname{clamp}(\cdot, \alpha)$ is to truncate the values into $[-\alpha, \alpha ]$.
We call this quantization as $L_p$ quantization where $p=\infty$ corresponds to the standard quantization methods.
As shown in Table~\ref{tab:optimizer_loss}, when training GPT on the FineWeb dataset, Lion converges poorly under $Q_\infty$ and 1-bit quantization, whereas $L_0$ and $L_1$ quantizations converge more effectively. Based on these findings, we will adopt $Q_1$ which uses the $L_1$ norm and is henceforth referred to as “L1 quantization” for all subsequent experiments.
This quantization method is simple yet sufficiently effective for practical applications, as will be demonstrated later. 
Its simplicity also results in low computational overhead.

\begin{figure}[t] 
  \centering
  \includegraphics[width=\linewidth]{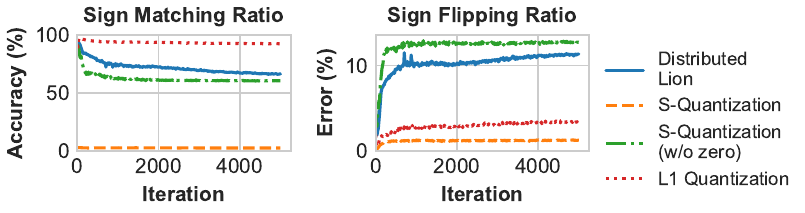}
  \caption{\textbf{L1 quantization approximates Lion well.}
    Top: Percent of update elements whose sign after quantization matches the standard Lion.
    Bottom: Percent of update elements whose sign after quantization reverses compared to standard Lion, changing the update direction. (This excludes elements mapped to 0.)}
  \label{fig:quant-error}
\end{figure}

\Cref{fig:quant-error} shows the accuracy of quantizing Lion updates with different methods in terms of whether the signs of the updates match or flip (introducing error).
Standard $Q_\infty$ stochastic quantization clips most values to zero, resulting in a low match and flip rate.
One approach to address this (``w/o zero'') is to stop rounding to zero and instead quantize sufficiently small values to $+1$ or $-1$.
While this improves the match rate, it does not match distributed Lion.
We also consider transforming values to a logarithmic scale before quantizing them, then reverting, which also does not match distributed Lion.
In contrast, our L1 quantization significantly improves the sign match rate and reduces the flip rate compared to distributed Lion.
We note that it may be surprising that distributed Lion is able to converge when its sign matching rate is 60--70\%, only a little better than random.
However, most of its errors are from returning zero, and consequently, its sign flip rate is around 10\%.
Further, the parameters that result in zeros often exhibit significant inter-worker variability, suggesting the need for updates in these cases is inherently low.

\begin{figure}[t]
  \centering
  \includegraphics[width=\linewidth]{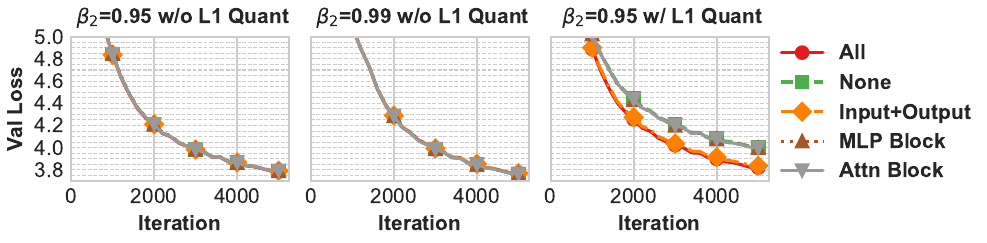}
  \caption{\textbf{(Left) Momentum synchronization or L1 quantization is necessary when $\beta_2$ is small.} 
  Momentum is never synchronized for “None”; all other settings synchronize momentum every 10 iterations. When $\beta_2$ is large, gradients are stable across workers, and synchronization is not needed. When $\beta_2$ is small, synchronization is necessary for key layers. 
  In addition, unlike standard distributed Lion, L1 quantization achieves convergence comparable to synchronized momentum, even without synchronization (“None”).}
  \label{fig:combined-quant}
\end{figure}


\section{On the necessity of momentum synchronization}

In standard Lion, while the final update vectors are 1-bit, the momentum is computed with full-precision gradients. In contrast, in distributed Lion~\citep{liu2024communication}, only the 1-bit updates are communicated, and the momentum term cannot be recovered, which can lead to momentum divergence between workers. Further, communicating the momentum would add overhead during training. We now investigate the extent to which momentum diverges and how to efficiently mitigate any divergence.

\subsection{Layerwise variance of momentum}


\begin{figure}[t]%
    \centering%
    \includegraphics[width=\linewidth]{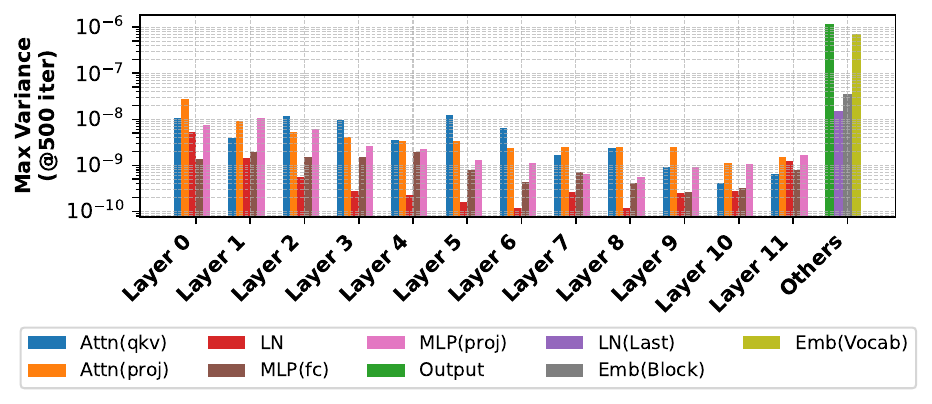}%
    \caption{\textbf{There is significant variance between workers for the first (embedding) and last layers.}
      We plot the maximum standard deviation between workers for each layer, observed for each element, while training GPT (730M) on OpenWebText.}\label{fig:var-layer}%
\end{figure}

\Cref{fig:combined-quant} compares multiple synchronization strategies for momentum: no synchronization, synchronizing every 10 iterations, and selective layer synchronization (only MLP or attention blocks).
When $\beta_2$ is large ($0.99$), we see that training converges similarly with all strategies; hence momentum synchronization is not required to maintain convergence.
In contrast, with a smaller $\beta_2$ ($0.95$), convergence deteriorates without synchronization.
Intuitively, this is due to a larger $\beta_2$ ensuring more stable gradients across iterations, which likely compensates for any differences in updates across workers.
\Cref{fig:variance-beta2} validates this, showing that momentum variance between workers decreases as $\beta_2$ increases.
Hence, the need for momentum synchronization depends primarily on the value of $\beta_2$ rather than on specific synchronization timing during training.

We can also see from \cref{fig:combined-quant} that synchronizing the momentum of only the input and final layers is sufficient to achieve convergence similar to that of synchronizing the momentum across all layers at $\beta_2 = 0.95$.
\Cref{fig:var-layer} compares the variance in momentum between workers, showing, for each layer, the maximum standard deviation measured for the element with the highest deviation across workers.
The standard deviation for the input embedding layer and the output layer is nearly 100 times larger than that of other layers.
This indicates that the momentum in these layers varies significantly between workers, whereas there is almost no variation in the other layers.
Given that the input and output layers are the layers where data directly enters, this result may be expected, considering that the data differs across workers.
In vision models, it is only the input layer that exhibits such a large standard deviation (see \cref{fig:var-res18}).
We leave a more detailed analysis of this to future research.

\subsection{Impact of quantization on momentum variance}


Our L1 quantization enables the communication of richer information than simple 1-bit updates, which we can also use to reduce the divergence in momentum between workers.
\Cref{fig:combined-quant} shows the results of training with 8 workers using L1 quantization to compress updates to the range $[-15, 15]$.
We are able to achieve similar convergence curves regardless of whether or not momentum is synchronized.

These findings suggest two approaches for good convergence: 1-bit communication with momentum synchronization or L1 quantization with no momentum synchronization.
The best choice depends on the specific setting.
As a general guideline, when there are few workers, using L1 quantization is recommended.
However, with a larger number of workers, the increased bitwidths required to correctly represent the majority vote limit potential speedups in training and quantization information.
For example, with 125 workers, a uint8 can can only support 0/1 summation and it becomes necessary to synchronize momentum occasionally.


\begin{figure}%
    \centering%
    \includegraphics[width=\linewidth]{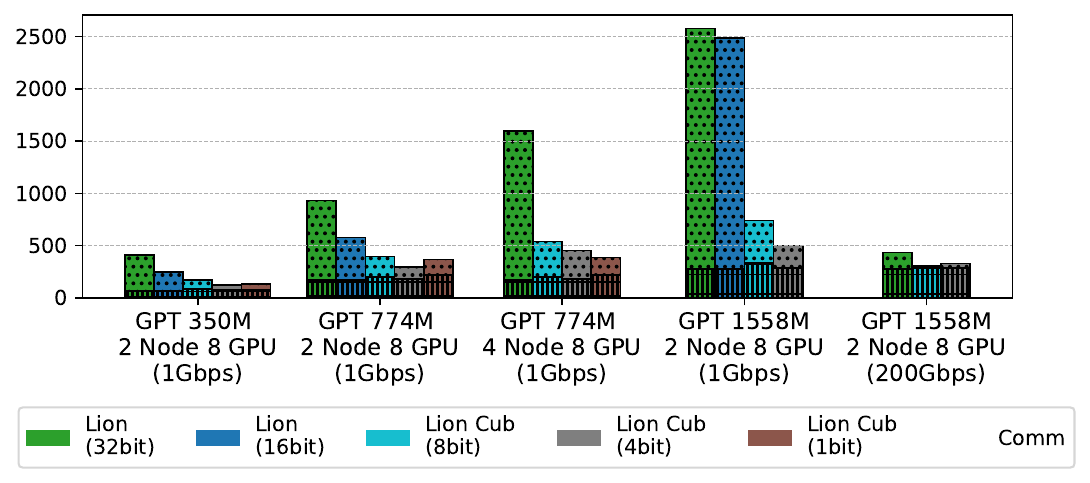}%
    \caption{\textbf{Breakdown of runtime performance when training GPT.}
      Lion Cub significantly reduces communication volume, leading to reduced communication time and faster end-to-end iterations.
      ``Comm'' includes both communication time and quantization overheads.}\label{fig:performance-model-gpt}%
\end{figure}

\section{Lion Cub}

We have presented a set of methods for accelerating communication and maintaining convergence while using the Lion optimizer, including 1-bit allreduces, $p$-bit allreduces with L1 quantization, and delayed momentum synchronization.
We refer to this collection of techniques for reducing Lion's communication as ``Lion Cub'', indicating a smaller, more communication-efficient version of Lion.


\subsection{Performance results}

We first consider the runtime performance of training.
\Cref{fig:performance-model-gpt} presents a breakdown of the runtime for training 350M, 774M, and 1558M parameter GPT models with standard Lion and Lion Cub in multiple precisions.
As communication precision decreases, communication time also decreases, in line with expectations.
This delivers up to 5.1\speedx{} speedups to end-to-end training over standard Lion in the lower-bandwidth setting.
Unfortunately, when training the 730M parameter model on 2 nodes, 1-bit Lion Cub in fact increases communication time; this is due to the overhead of packing and unpacking the compressed representation outweighing the decreased communication time.
However, if we increase the number of nodes, this only increases the communication time, not the pack/unpack time, resulting in further communication reductions.
As model size increases from 350M into 1558M, the communication bottleneck intensifies, yielding greater performance improvements for Lion Cub over standard Lion.
Even in the higher-bandwidth setting, Lion Cub reduces iteration time compared to standard Lion.

\subsection{Training results}

\begin{figure}[t]%
    \centering%
    \includegraphics[width=\linewidth]{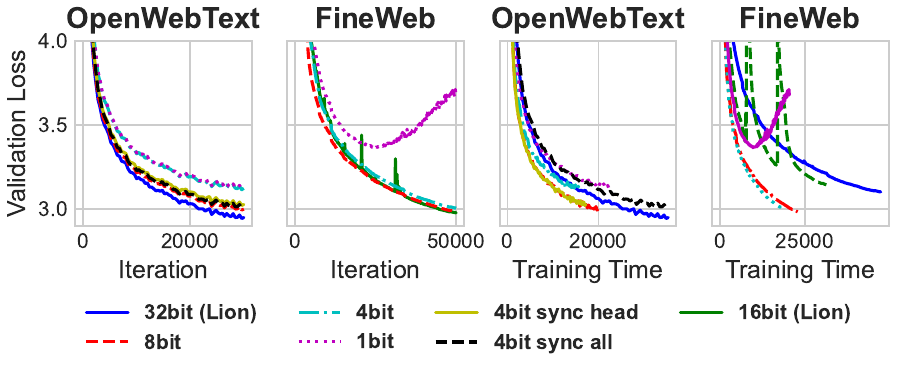}%
    \caption{\textbf{Validation loss with GPT (774M)} when training with 8 workers on 2 nodes in our Ethernet configuration.
      Lion Cub reduces training time compared to standard Lion and can maintain comparable convergence.
      Note in 4-bit Lion Cub, workers only send 1/-1 values, whereas in 8-bit Lion Cub, we use L1 quantization and send values in $[-15, 15]$.
      In 4-bit, sync head Lion Cub, we synchronize the momentum of the final layer every 10 iterations.
      Sync all synchronizes the momentum of all layers.}\label{fig:train_curve_tsubame_774}%
\end{figure}

We now validate that Lion Cub preserves convergence and model quality.
\Cref{fig:train_curve_tsubame_774} presents training curves for GPT (774M) trained on the OpenWebText dataset and Fineweb dataset with 8 workers on 2 nodes.
Lion Cub (8-bit) maintains convergence comparable to the 32-bit Lion, while Lion Cub (4-bit) shows a decline in convergence.
Consequently, Lion Cub (8-bit) with L1 quantization achieves faster real-time convergence.
The convergence quality of Lion Cub (4-bit) can be recovered by synchronizing the momentum of the head layer every ten iterations, without significantly impacting training time.
On the other hand, synchronizing momentum across all layers creates a communication bottleneck, which limits the speedup in real-time training performance.
We trained on the Fineweb dataset with more aggressive hyperparameter settings than those used for the OpenWebText dataset.
Under these conditions, LionCub(1-bit) failed to converge, forcing us to adopt LionCub(4-bit). Because this setup is highly sensitive to the numerical precision, compressing gradients from 32-bit to 16-bit (bfloat16) prior to communication (the 16-bit Lion) resulted in unstable training. 
Overall, Lion Cub, employing L1 quantization, provides substantial benefits in both training stability and communication speed.

\begin{figure}[t]%
    \centering%
    \includegraphics[width=\linewidth]{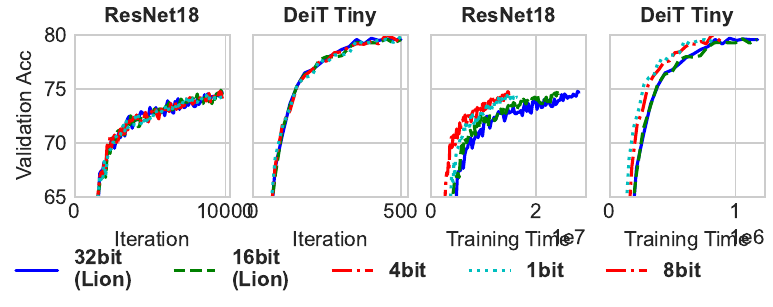}%
    \caption{\textbf{Validation accuracy for training ResNet-18 on CIFAR100 and fine-tuning DeiT-Tiny on CIFAR100} using 8 workers on 2 nodes in our lower-bandwidth configuration. Interestingly, if you look at the training curve per iteration, even 1-bit allreduce closely mirrors Lion with 32-bit precision.}\label{fig:resnet18-cifar10}%
\end{figure}

We also evaluate Lion Cub in two additional contexts for vision tasks in \cref{fig:resnet18-cifar10}.
The first task is training ResNet-18 on CIFAR100 for 100 epochs.
Second, we fine-tune a DeiT-Tiny model, pretrained on ImageNet, on CIFAR100 for 20 epochs.
In both cases, Lion Cub achieves comparable convergence to Lion in less overall time.
Further, this is achieved even when not using L1 quantization or momentum synchronization.
Hence, it is possible to lower the communication volume significantly to reduce overall training time in these cases.

\section{Discussion and Conclusion}

\paragraph{Limitations}
Distributed training methods, especially ones incorporating compression, are notoriously fickle, and their quality is often sensitive to the scale of the model and cluster being used.
While we have demonstrated Lion Cub’s strong performance at a relatively small scale, computational resource limitations have so far prevented confirmation of its benefits at larger scales.
We also note that we have focused on communication in a data-parallel setting, as opposed to model-parallelism techniques.
While we believe Lion Cub could be extended to these approaches, we leave this to future work.

Another significant limitation is that while our method performs very well in lower-bandwidth networks, the improvements are less dramatic in low-latency, high-bandwidth networks.
Despite this, we believe that distributed training in lower-bandwidth environments remains essential.
We expect communication reduction to become increasingly important in the future.
As deep learning becomes democratized, many more people will be training models on clusters with relatively low-end interconnects and will be communication-bound without improved techniques like Lion Cub.
But even on high-end clusters, trends in hardware indicate that the balance of communication and computation will continue to worsen as GPU performance improves faster than network bandwidth, necessitating reduced communication for scalability and efficiency.

\paragraph{Conclusion}
We presented Lion Cub, a suite of improvements enabling reduced communication overhead for distributed training with Lion.
Lion Cub incorporates optimized communication primitives and a novel L1 quantization method, enabling it to converge comparably to full-precision Lion on a variety of models, including ResNets and GPT, while reducing overall runtime.


\bibliography{ref}
\bibliographystyle{icml2025}

\newpage
\appendix
\onecolumn
\section*{Acknowledgements}

This work was performed under the auspices of the U.S. Department of Energy by Lawrence Livermore National Laboratory under Contract DE-AC52-07NA27344 (LLNL-TR-871246). T.B.N., B.V.E., and N.D. were funded by LLNL LDRD \#24-SI-008.
This work used computational resources TSUBAME4.0 supercomputer provided by Institute of Science Tokyo through the HPCI System Research Project (Project ID: hp240170)
This work is supported by JST CREST Grant Number JPMJCR2112.

\section{Extended Related work}
\subsection{SignSGD}
SignSGD is an optimization method that applies the $\sign$ operator to gradients, resulting in three possible values: +1, 0, and -1~\citep{seide20141,bernstein2018signsgd,bernstein2018signsgdmaj}.
This approach has gained attention due to its similarity to Adam, particularly in terms of the normalized update~\citep{kunstner2023noise}, but also as a distributed learning method when combined with majority voting, as updates can be compressed~\citep{bernstein2018signsgdmaj}.
In majority voting, $\sign$ is first applied to the gradients, and then again to their sum: $\theta_{t+1}=\theta_t-\eta_t \sign\left[\sum_{i=1}^N \sign \nabla_{\theta} f_i(\theta_{t-1}) \right]. $
Note that $\sign$ is unbiased, which is why a stochastic $\sign$ is sometimes employed~\citep{safaryan2021stochastic}.
Typically, discussions of majority voting assume a parameter server model for communication, where the signs computed on each worker are sent to a central server which computes the global sign and returns the result.
As we will discuss, while straightforward, this approach does not scale.

\subsection{Heavy-tailed noise and Gradient Clipping}

We can regard our L1 quantization as performing quantization after gradient clipping. 
In heavy-tailed centralized training, gradient clipping is known to be an effective approach to mitigate heavy-tailed noise~\citep{gorbunov2020stochastic, zhang2020adaptive}.
In distributed optimization, several works have shown that clipping gradients before communication helps curb the influence of heavy-tailed noise~\citep{yang2022taming,xiang2023betastochastic,yan2024improved}.
FAT-Clipping integrates gradient clipping into federated averaging, either once per round or at each local step, to tame fat-tailed noise and guarantee convergence~\citep{yang2022taming}.
Similarly, EPISODE applies episodic gradient clipping, which uses the global gradient norm at the start of each round to decide whether to clip local updates, to adaptively bound heavy-tailed stochastic noise and ensure robust convergence under data heterogeneity~\citep{crawshaw2023episode}.
Furthermore, TQSGD first applies gradient clipping and then quantizes the clipped gradients for communication, thereby compensating for heavy-tailed noise, and is the method most closely related to our L1 quantization~\citep{yan2024improved}. 
In contrast to these methods, L1 quantization automatically performs both scaling and clipping without any additional hyperparameters, and further combines this mechanism with the Lion optimizer. Although we do not provide formal convergence guarantees, our empirical evaluation offers a practical demonstration of the importance of clipping before quantization in large-scale training.

\subsection{Layerwise Selective Communication}

We synchronize the momentum of selected layers every few iterations. Although no prior work has performed layer-wise selective momentum synchronization, federated learning has long cut communication costs by transmitting only certain layers’ weights or gradients more frequently or with higher precision. 
TWAFL is an asynchronous federated learning scheme in which the parameters of deep layers are updated less often than those of shallow layers~\citep{chen2020communication}. 
Likewise, FedLAMA reduces communication overhead by dynamically choosing each layer’s optimal aggregation interval by evaluating model discrepancy and communication costs~\citep{lee2023layer}.
TLAQC focuses on quantizing and communicating only two representative layers of the model, while all other layers merely accumulate updates locally~\citep{ren2023two}. 
In FedPart, each client selectively updates and transmits only a subset of the network (for example, just the feature‐extraction layers or only the classification layers) each round~\citep{wang2024why}. 
Thus, in the federated‐learning, it has long been known that different layers exhibit different degrees of importance to synchronize. 
Although our approach keeps weights and gradients fully synchronized across all GPUs, letting only the momentum terms diverge, we have found that the amount of momentum synchronization required varies from layer to layer.

\section{Communication Design}
\subsection{The impact of 1-bit allreduces}


In both allreduce implementations, all updates are compressed to 1 bit.
However, this loses information, since the $\sign$ operator is ternary and we no longer represent values which are exactly 0.
We next examine the impact this has on training, focusing on the 1-bit compressed allreduce, which performs all communication in 1-bit precision.

The first allreduce stage uses an all-to-all to transmit update vectors.
True zeros are rare in these, as a gradient must be consistently zero over many steps due to the momentum.
One situation where this does occur is when training transformers, where certain parameters in the embedding layer could output zeros if the sequence length is always shorter than the model's maximum supported sequence length.
The error introduced by 1-bit compression can accumulate over time in these parameters.
However, this situation can be easily identified in advance and the updates masked.

\begin{figure}[t] 
  \centering
  \includegraphics[width=0.7\linewidth]{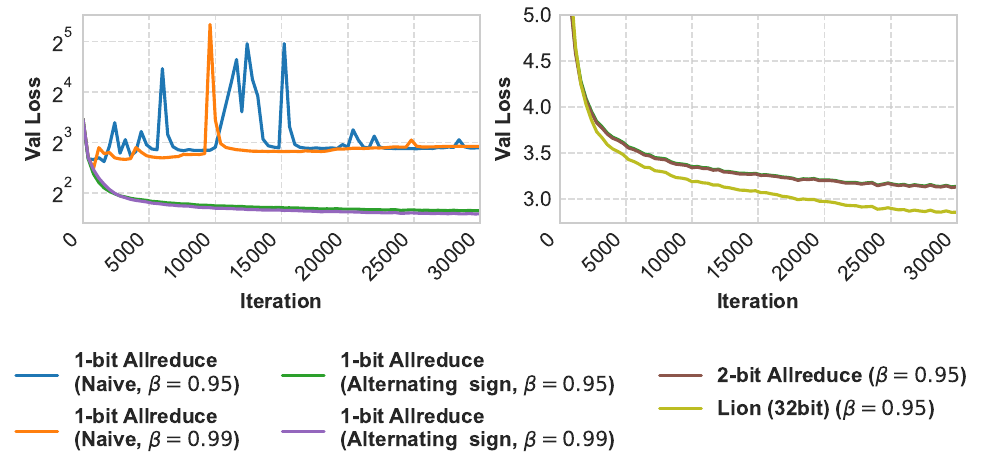}
  \caption{\textbf{1-bit allreduces, unable to represent exact zeros, diverge without adjustments.}
    The naive implementation of Lion using a 1-bit allreduce, which does not transmit exact zeros, leads to error accumulation and causes the loss to diverge.
    However, when zeros are converted to 1 or -1 on alternating iterations, the loss remains stable, regardless of the $\beta_2$ value.}
  \label{fig:loss-spiking}
\end{figure}

In the second stage, an allgather disseminates aggregated updates, also with 1 bit.
This is more challenging, because, in the typical case when there are an even number of workers, there can be an equal number of 1 and -1 updates for a parameter among the workers.
Should this occur, the conversion to 1 bit can change the descent direction, impacting convergence.
Standard Lion does not face this issue, as it applies the $\sign$ after summing updates.
We analyzed the update vectors produced by standard Lion and the 1-bit allreduce when training a GPT model in \cref{tab:misclassification}.
With 4 workers, over 30\% of the parameters have an update where the majority vote is tied, leading to an erroneous update depending on how it is transmitted.
While this rate decreases with 8 workers (as ties are less likely), it is still significant.
Indeed, as \cref{fig:loss-spiking} demonstrates, these errors can lead to loss spikes and training divergence, and so are necessary to address, even if they may be relatively unlikely at scale.
We find that a simple alternating sign method is sufficient to address this: we convert 0 to 1 on odd iterations and to -1 on even iterations before compressing to 1 bit.
We considered a stochastic approach, but did not proceed with it as the overhead of random number generation was significant.
\Cref{fig:loss-spiking} shows that the alternating sign approach successfully stabilizes training and has minimal differences to using a 2-bit allreduce which transmits exact zeros.


\begin{SCtable}[2][t]
  \centering
  \scriptsize
  \renewcommand{\arraystretch}{1.2}
  \setlength{\tabcolsep}{5pt}
  \begin{tabular}{lcccc}
    \toprule
    & & \multicolumn{3}{c}{$\sign(\textrm{majority vote})$} \\
    \cmidrule{3-5}
    & & \textbf{1} & \textbf{-1} & \textbf{0} \\
    \cmidrule{2-5}
    \multirow{4}{*}[2ex]{\rotatebox{90}{\shortstack{Standard\\Lion (FP32)}}}
      & \textbf{1}  & $29.15\%/32.08\%$ & $2.26\%/4.34\%$ & $18.59\%/13.54\%$ \\
      & \textbf{-1} & $2.26\%/4.34\%$  & $29.14\%/32.13\%$ & $18.59\%/13.56\%$\\
      & \textbf{0}  & $0\%/0\%$        & $0\%/0\%$        & $0\%/0\%$       \\
    \bottomrule
  \end{tabular}
  \caption[\textbf{Update values under majority vote vs standard Lion.}]{%
    \textbf{Update values under majority vote vs standard Lion.}
    This shows the percent of update values with each sign under majority voting versus standard Lion from the 3000th iteration of training a GPT (730M) model.(Entries are for 4 / 8 workers.)
  }
  \label{tab:misclassification}
\end{SCtable}

\section{Additional Explanation for Quantization}
\subsection{Toy Experiments}

\begin{figure}
    \centering
    \includegraphics[width=\linewidth]{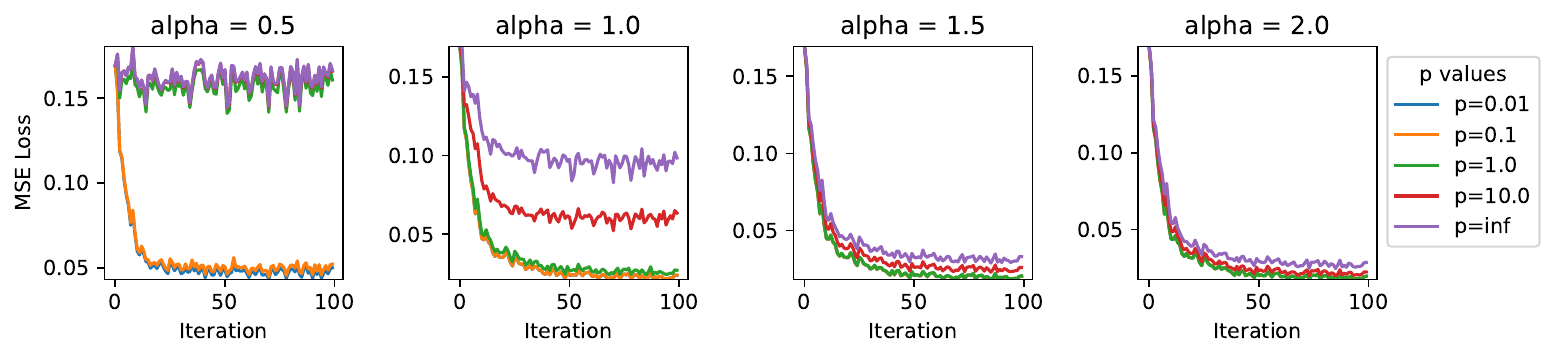}
    \caption{\textbf{Smaller $p$ values enhance $L_p$Quantization under heavy-tailed noise.}}
    \label{fig:toy-quant-error}
\end{figure}

In Fig. \ref{fig:toy-quant-error}, we implement a simple teacher–student MLP on synthetic data and introduce per‐client heavy‐tailed gradient noise sampled via the alpha stable Levy distribution (where $\alpha=0.5$ recovers symmetric Levy noise and $\alpha=2$ recovers Gaussian noise).  
At each iteration, the noisy gradients from $C=8$ clients are quantized by an $L_{p}$ quantizer for $p\in\{10^{-2},10^{-1},1,10,\infty\}$ (the case $p=\infty$ 
corresponds to the standard direct uniform quantizer).    
Figure \ref{fig:toy-quant-error} compares the baseline direct quantizer ($p=\infty$) against the family of $L_{p}$ quantizers under varying $\alpha$.
As these results demonstrate, the more heavy‐tailed the gradient noise becomes, the more the performance of the commonly used $L_{\infty}$ quantizer degrades, widening the gap with $L_{0}$ quantization.  
This suggests that if gradient noise is heavy‐tailed such as Transformer training or language model training, an appropriate $L_{p}$ quantization (with smaller p) may be especially important.

\subsection{Additional Experiments with vision models}

\begin{SCfigure}[0.55][t] 
  \centering
  \includegraphics[width=0.55\textwidth]{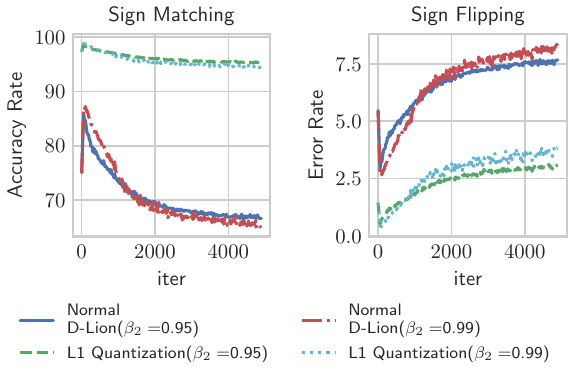}
  \caption{\textbf{Applying L1 quantization in vision models also improves the alignment rate.}
    We measured the same metrics as in Figure~\ref{fig:quant-error} for ResNet18 on CIFAR100, and found that L1 quantization similarly increases the alignment ratio in a vision model. This training was conducted with 8 GPUs. The variance in the input layer is significantly larger than that of other layers.}
  \label{fig:quant-resnet-element-acc}
\end{SCfigure}

\paragraph{Error rate}
Figure~\ref{fig:var-res18} shows that the effects of L1 quantization observed in a vision setting, specifically during ResNet18 training on CIFAR100, are similar to those observed in Figure~\ref{fig:quant-error}.
The results indicate that L1 quantization enables a close approximation of Lion in the ResNet18 vision model.
Interestingly, while the approximation to Lion worsens without L1 quantization, as shown in Figure~\ref{fig:resnet18-cifar10}, the 1-bit allreduce method still performs well.
This suggests that in this particular setup, effective optimization is possible without closely approximating Lion. 
Further investigation of this observation is left to future work.

\begin{SCfigure}[1.5][t]
    \centering
    \includegraphics[width=0.5\linewidth]{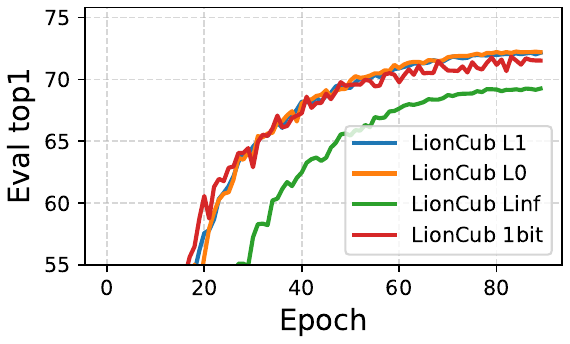}
    \caption{\textbf{ImageNet ablation study of quantization function.}
    We train ResNet-50 on ImageNet using different quantization functions. The training runs on 16 workers with a total batch size of 16,384. Because we use large-batch training, the final accuracy is lower than that of the standard 90-epoch ImageNet training.}
    \label{fig:imagenet_train_curve}
\end{SCfigure}

\paragraph{ImageNet}

As shown in Table~\ref{fig:update_distribution}, $L_1$ quantization yields the lowest training loss during GPT pre-training, while both $L_0$ and $L_\infty$ quantizations incur higher losses. 
To verify whether this behavior extends to vision models, we trained ResNet-50 on ImageNet for 90 epochs (Figure~\ref{fig:imagenet_train_curve}). 
The results reveal that $L_0$ quantization achieves accuracy on par with, or slightly above, $L_1$ quantization, making it the best-performing quantization scheme overall. 
Although $L_1$ quantization remains a viable alternative due to its comparable accuracy, $L_0$ quantization demonstrates the highest top-1 performance. 
These observations are consistent with the toy experiments in Figure~\ref{fig:toy-quant-error}. 
A more detailed analysis of the accuracy–efficiency trade-off between $L_0$ and $L_1$ quantization is left for future work.

\paragraph{CIFAR5m}

\begin{figure}
    \centering
    \includegraphics[width=\linewidth]{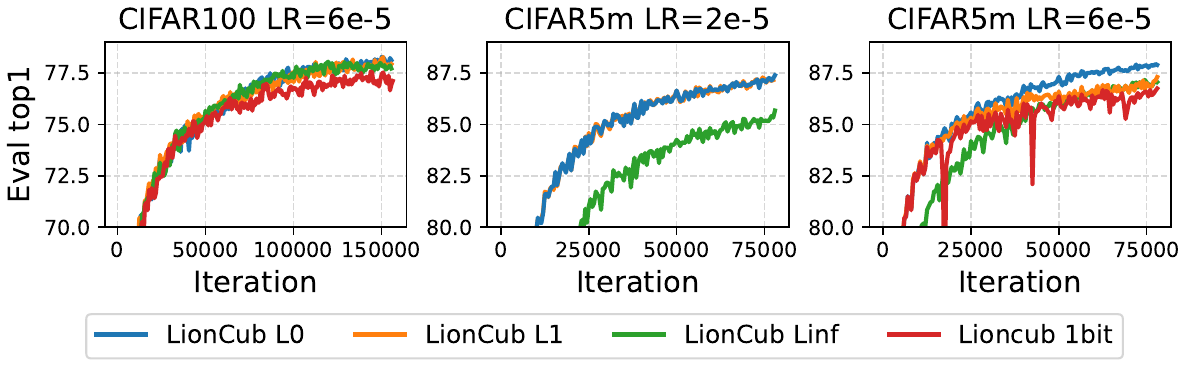}
    \caption{\textbf{CIFAR ablation study of quantization function.}
    Accuracy of ResNet-18 on CIFAR100 and CIFAR5m using $L_\infty$, $L_0$, and $L_1$ quantization. While all three schemes yield nearly identical results on CIFAR100, the choice of quantization method produces clear performance differences on CIFAR5m.}
    \label{fig:cifar-quantization}
\end{figure}

Figure~\ref{fig:cifar-quantization} presents an ablation study of quantization functions for ResNet-18 trained on CIFAR5m and CIFAR100. 
Note that CIFAR-5m consists of six million synthetic, CIFAR-10-style images generated by a DDPM-based model\footnote{\url{https://github.com/preetum/cifar5m}}.
On CIFAR100, model accuracy is largely insensitive to the choice of quantizer: $L_\infty$, $L_0$, and $L_1$ quantization achieve nearly identical performance. Remarkably, this result mirrors the behavior observed in our toy experiment (Figure~\ref{fig:toy-quant-error}), where the weight distributions on CIFAR100 appeared not to be heavy-tailed but rather approximated Gaussian or uniform distributions.
By contrast, when training on CIFAR5m with a learning rate of $2\times10^{-5}$, accuracy clearly depends on the quantization scheme: $L_\infty$ quantization underperforms compared to $L_0$ and $L_1$. Intriguingly, increasing the learning rate to $6\times10^{-5}$ causes $L_0$ quantization to surpass both the standard Lion optimizer and the other quantizers in accuracy. We hypothesize that the $L_0$ constraint acts as an implicit clipping mechanism, stabilizing weight updates under higher learning rates.
These contrasting trends between the online-style CIFAR5m and the multi-epoch CIFAR100 highlight an important consideration: for developing and evaluating training methods for large-scale tasks—such as ImageNet classification or GPT language modeling—CIFAR5m may serve as a more representative benchmark than CIFAR100.

\section{Additional Training Experiments}
\subsection{Additional Explanation for Momentum Synchronization}

\begin{figure*}[t]%
    \centering%
    \includegraphics[width=\linewidth]{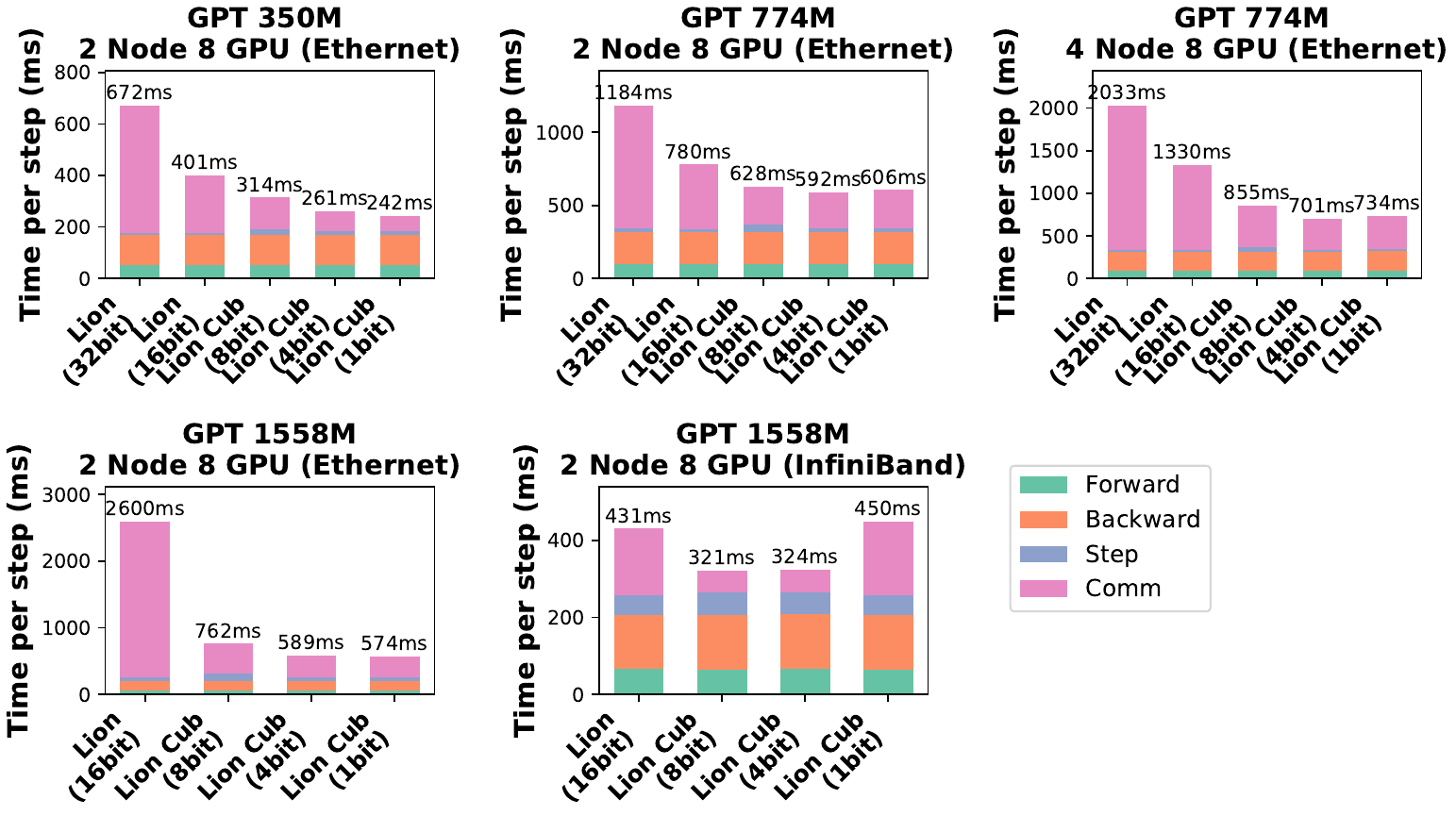}%
    \caption{\textbf{Breakdown of runtime performance when training GPT.}
      Lion Cub significantly reduces communication volume, leading to reduced communication time and faster end-to-end iterations.
      ``Comm'' includes both communication time and quantization overheads.}\label{fig:performance-model-gpt-detailed}%
\end{figure*}

\begin{SCfigure}[0.8][t]
    \centering
    \includegraphics[width=0.6\linewidth]{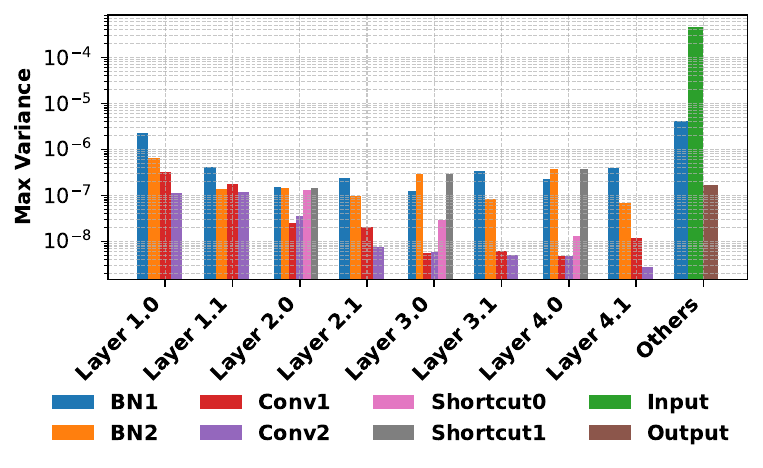}
    \caption{\textbf{The variance between workers is particularly large in the input layer.}
    We trained resnet18 on CIFAR100. }
    \label{fig:var-res18}
\end{SCfigure}

\begin{SCfigure}[1.5][t]
  \centering
  \includegraphics[width=0.3\linewidth]{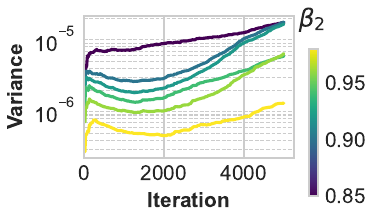}
  \caption{
  \textbf{The larger the $\beta_2$, the smaller the variance between workers.}
    We plot the progression of the maximum standard deviation between workers, which is observed for each element.
    The standard deviation decreases as $\beta_2$ increases.
    Additionally, the standard deviation tends to remain consistent throughout training.}
  \label{fig:variance-beta2}
\end{SCfigure}

\paragraph{Vision Model}
The results of similar observations from Figure~\ref{fig:var-layer} are presented in Figure~\ref{fig:var-res18}, showing the variance in momentum across layers for ResNet18 trained on CIFAR100. In the case of ResNet18, the variance was particularly high in the input layer. Unlike in GPT training, where both input and output layers exhibited high variance, it is interesting that only the input layer shows such variance in the vision model, with little variance observed in the output layer.
In vision models, the smaller output dimension of the output layer may influence this variance, though further investigation is left for future work.

\subsection{Additional experiments with language models}
\label{sec:appendix-gpt-exp}
\begin{SCfigure}[1.5][t]
  \includegraphics[width=0.5\textwidth]{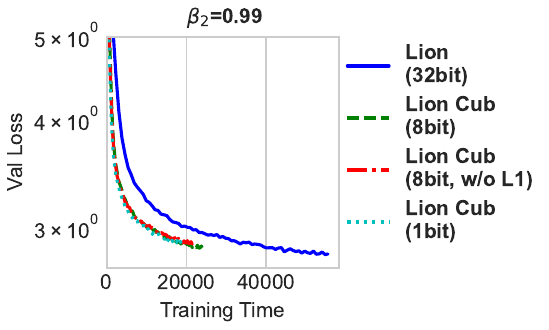}
  \caption{\textbf{Validation loss on OpenWebText with GPT-2 (735MB) when training with 4 nodes.}
  We trained a 735M parameter GPT model on OpenWebText with 16 GPUs using a cluster equipped with 4 H100 GPUs per node.
  In a 4-node training setup, we can not use the 4-bit version of Lion Cub because it exceeds the representable value range. Thus, we must use either the 8-bit or 1-bit variant of Lion Cub.}
  \label{fig:gpt-735m-4node}
\end{SCfigure}

\begin{figure}[htb]
  \centering
  \begin{subfigure}[b]{0.49\linewidth}
    \centering
    \includegraphics[width=\linewidth]{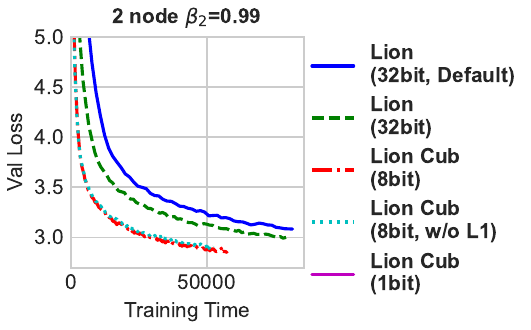}
    \caption{$n=2$}
    \label{fig:val-loss-n2}
  \end{subfigure}%
  \hfill
  \begin{subfigure}[b]{0.49\linewidth}
    \centering
    \includegraphics[width=\linewidth]{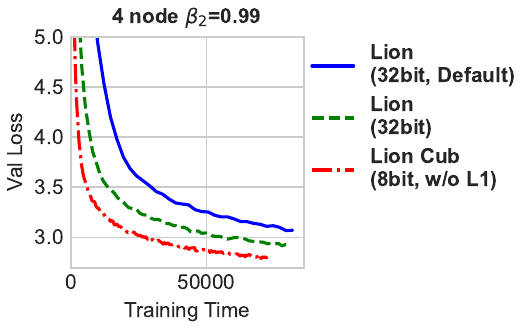}
    \caption{$n=4$}
    \label{fig:val-loss-n4}
  \end{subfigure}%
   \caption{\textbf{Validation loss on OpenwebText with GPT-2(1.5B).}
    We trained a 1.5B parameter GPT model on OpenWebText using a cluster equipped with 4 H100 GPUs per node. 
    With low-latency environments, this approach significantly reduces training time.}
  \label{fig:gpt1b-training-curve}
\end{figure}

We conducted further experiments on GPT training by varying the number of nodes and model sizes, as shown in Figure hoge. In Figure \ref{fig:gpt-735m-4node}, a 735M GPT model is trained across 4 nodes. Although 4-bit allreduce cannot be used in this case, we still observe a notable speedup over standard Lion. Additionally, Figure \ref{fig:gpt1b-training-curve} presents training results for a 1.5B model. As indicated by this graph, reducing training time with Lion Cub becomes increasingly crucial as model size grows to 1.5B.
In Figure~\ref{fig:gpt1b-training-curve}, “Default” refers to the standard 32-bit Lion implementation, where training was conducted using PyTorch’s DDP module without modifications. We observed that combining gradients into a single tensor and performing a single allreduce operation was faster than PyTorch’s default approach. Consequently, for all experiments presented in this paper, we consolidated gradients into one tensor and executed only one allreduce call per iteration.

\begin{SCfigure}[1.5][t]
  \includegraphics[width=0.6\textwidth]{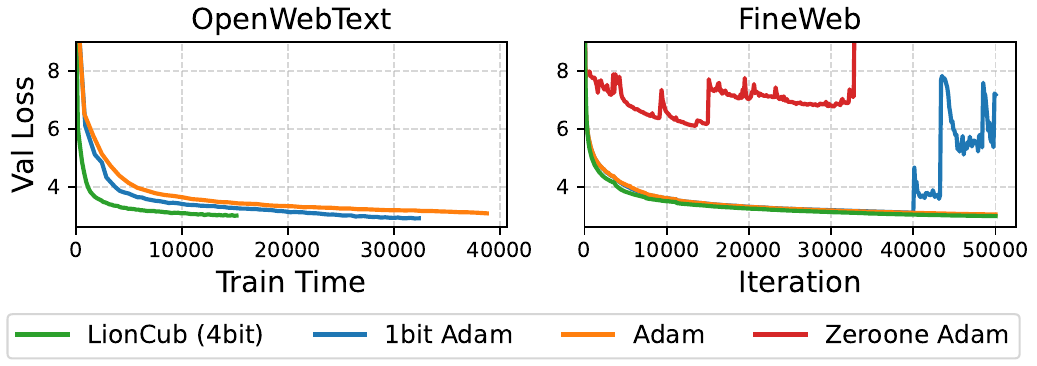}
  \caption{\textbf{Comparison with Adam.}
  We trained a 735M GPT model on OpenWebText and FineWeb with 8 GPUs.
  }
  \label{fig:adam_variants}
\end{SCfigure}

In Figure \ref{fig:adam_variants}, we compare LionCub with communication efficient variants of Adam. 
When training on OpenWebText under restricted bandwidth, the 1-bit Adam outperforms standard Adam in terms of total training time; however, because LionCub has even lower communication and compression overhead, it achieves the fastest overall training.
On the FineWeb corpus, using the same hyperparameters that lead Adam to converge causes both 1-bit Adam and 0/1 Adam to diverge. 
It may be possible to restore stable convergence for those variants through further tuning of learning rates or other settings.
Moreover, 1-bit Adam must run for roughly 23000 iterations before freezing its momentum term by default, at which point its communication cost drops. 
In training regimes with only a few thousand iterations, LionCub, which does not require any warm-up phase, can achive more stable and rapid training.
However, a comprehensive assessment of Adam’s communication-reduction techniques versus LionCub will require a more fine-grained analysis of overheads, convergence behavior, and hyperparameter sensitivity and it is left for future work.

\section{Additional profiling results}
\subsection{Different network bandwidth}
\begin{table}[t]
  \centering
  \caption{\textbf{Total training time and communication overhead ratio for different network speed.}}
  \label{tab:training_time_different}
  \sisetup{
    detect-mode,
    table-number-alignment = center,
    round-mode = places,
    round-precision = 1,
    separate-uncertainty = false
  }
  \begin{tabular}{
    l
    *{3}{ S[table-format=2.1]  S[table-format=2.1] }
  }
    \toprule
    & \multicolumn{2}{c}{\textbf{1\,Gbps}}
    & \multicolumn{2}{c}{\textbf{10\,Gbps}}
    & \multicolumn{2}{c}{\textbf{200\,Gbps}} \\
    \cmidrule(lr){2-3} \cmidrule(lr){4-5} \cmidrule(lr){6-7}
    \textbf{Optimizer}
      & {\textbf{1step (s)}}
      & {\textbf{Comm.\ (\%)}}
      & {\textbf{1step (s)}}
      & {\textbf{Comm.\ (\%)}}
      & {\textbf{1step (s)}}
      & {\textbf{Comm.\ (\%)}} \\
    \midrule
    Lion(32bit)     & 0.97 & 74.8 & 0.45 &  43.6 &  0.33 &  22.0 \\
    Lion(16bit)    & 0.64 & 60.3 &  0.36 &  30.3 &  0.31 &  14.7 \\
    LionCub(8bit)  & 0.48 & 41.8 &  0.34 &  20.8 &  0.30 &  9.7 \\
    LionCub(4bit)  & 0.32 & 32.2 &  0.31 &  19.3 &  0.30 &  11.8 \\
    LionCub(1bit)  & 0.42 & 41.1 &  0.35 &  28.1 &  0.33 &  25.0 \\
    \bottomrule
  \end{tabular}
\end{table}

Using NCCL\_IB\_HCA, we measured at three different communication bandwidth levels within the same cluster. The results are presented in Figure \ref{tab:training_time_different}. 
The 1-bit LionCub implementation uses an AlltoAll algorithm, which in some cases yields lower throughput than the 4-bit (and higher) LionCub variants that use AllReduce. 
Note that when a high-speed interconnect (e.g. 200 Gbps InfiniBand) is available, the cost of quantization and other computations can dominate overall performance.
Conversely, when limited to 1 Gbps Ethernet, the 32-bit Lion implementation spends 74 \% of its time on communication; using to the 4-bit LionCub cuts this share to nearly half.

\subsection{Memory} 

\begin{SCfigure}[1.5][h]
  \includegraphics[width=0.6\textwidth]{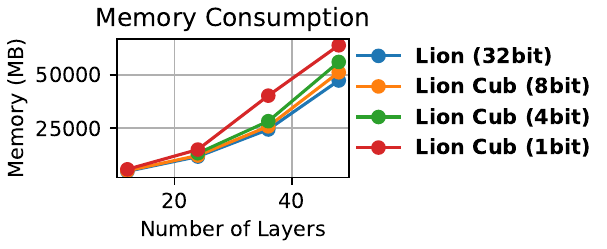}
  \caption{\textbf{Maximum memory consumption when training GPT using Lion Cub.}
  This figure illustrates the peak memory usage across different layers during GPT training with the Lion Cub optimizer.}
  \label{fig:memory-vs-layer}
\end{SCfigure}

We measured the memory consumption for training GPT with Lion Cub in Figure~\ref{fig:memory-vs-layer}. For smaller model sizes, memory consumption is nearly the same across all methods. However, as model size increases, the memory usage of the Lion cub with 1-bit allreduce becomes higher than the others. This is due to the substantial, unavoidable temporary memory required for packing and unpacking data to a 1-bit format.

\subsection{Vision model} 

\begin{figure*}[t]%
    \centering%
    \includegraphics[width=0.8\linewidth]{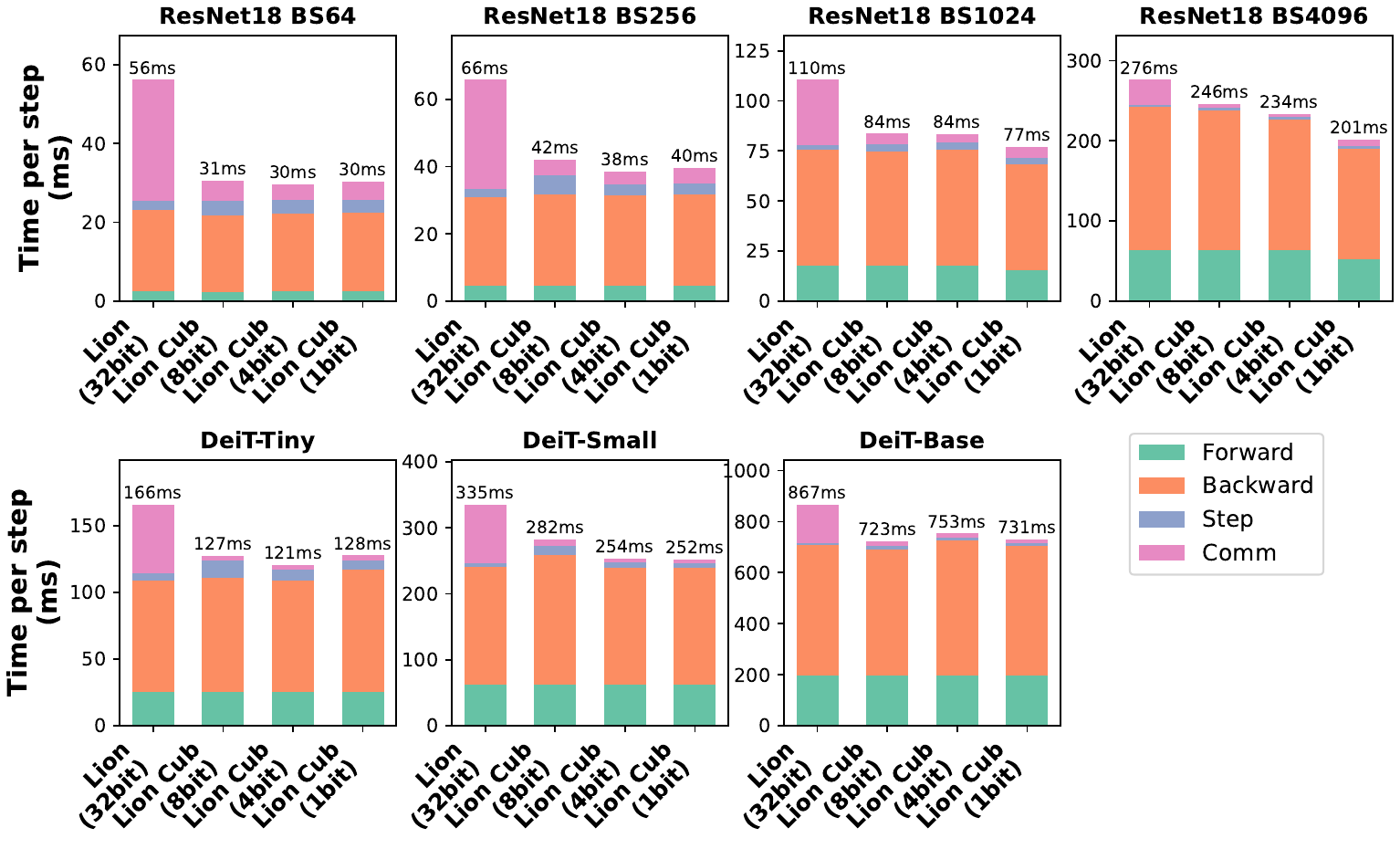}%
    \caption{\textbf{Breakdown of runtime performance when training ResNet and DeiT.}
      All results use 2 nodes and our ``Ethernet'' configuration and ``Comm'' includes both communication time and quantization overheads.
      Lion Cub significantly reduces communication and overall iteration time compared to Lion.}\label{fig:performance-model-resnet}%
\end{figure*}

We present results with ResNet-18~\citep{he2016deep} and DeiT-Tiny, -Small, and -Base~\citep{touvron2021training} on CIFAR100~\citep{krizhevsky2009learning} in \cref{fig:performance-model-resnet}, where Lion Cub similarly reduces both communication and training time.
Here we also study the impact of the batch size: as batches grow, computation increases while communication remains constant.
We see this effect in \cref{fig:performance-model-resnet}, where the benefits of Lion Cub are more modest with large batch sizes.
Hence, we expect that in large-batch training regimes with smaller models, as is common in vision tasks, reducing communication is of limited benefit.
However, it has recently become a trend to train very large vision models, where communication reduction is more critical.

\section{Experimental settings}
\label{sec:app-experimental-settings}

\subsection{Implementation Details}

\begin{algorithm}[thb]%
\caption{Distributed Lion with Quantization}\label{alg:lion_optimizer}%
\begin{algorithmic}[1]
\STATE \textbf{Given} $\beta_1, \beta_2, \lambda, \eta, f, Q$;\quad \textbf{Initialize} $\theta_0, m_0 \gets 0$
\WHILE{$\theta_t$ not converged}
    \STATE $c_{i,t} \gets \beta_1 m_{i, t-1} + (1 - \beta_1) \nabla_\theta f_i (\theta_{t-1})$
    \STATE $c_{t}^{*} \gets \sum_{i=1}^N Q(c_{i,t})$  (\text{sync update vector})
    \STATE $\theta_t \gets \theta_{t-1} - \eta_t \cdot \text{sign}(c_{t}^{*}) + \lambda \theta_{t-1}$
    \STATE $m_{i,t} \gets \beta_2 m_{i, t-1} + (1 - \beta_2) \nabla_\theta f_i (\theta_{t-1})$
\ENDWHILE
\end{algorithmic}%
\end{algorithm}

We implement Lion Cub in PyTorch~\citep{paszke2019pytorch} 2.2.0 and use NCCL~\citep{nccl} for communication.
For our 1-bit allreduce, we built upon the DeepSpeed \texttt{compressed\_allreduce} implementation~\citep{tang21a} and adapted it to Lion.
We identified the packing and unpacking of the 1-bit representation as a bottleneck that can dominate communication time, and implemented optimized routines using CuPy~\citep{nishino2017cupy}.

We conduct all experiments on a cluster where each compute node has four NVIDIA H100 GPUs with 94 GB of memory.
Nodes are interconnected with 4$\times$ NDR200 200 Gbps InfiniBand NICs.
To simulate a slower Ethernet network, we set the \texttt{NCCL\_IB\_DISABLE} environment variable to prevent NCCL from using the InfiniBand transport, and instead fall back to sockets with significantly lower bandwidth.

For our $p$-bit allreduce, we directly use NCCL's allreduce with its standard summation reduction.
We select the datatype for communication based on $p$ and the number of processes communicating.
When fewer than 8 bits are needed, we pack entries into a uint8; however, we currently require the bitwidth to evenly divide the size of the datatype we use.
To maximize efficiency, we map $\{-1, 1\} \mapsto \{0, 1\}$, which saves one bit per value.
For example, when performing a 4-bit allreduce among 8 workers, we simply sum in 0/1 format.
For an 8-bit allreduce among 8 workers, we apply L1 quantization to map values to $[0, 15]$ before allreducing.

We note that the 1-bit compressed allreduce uses an all-to-all to implement a reduce-scatter operation.
However, this operation cannot be directly implemented with the standard reduce-scatter in NCCL or MPI due to the need to change bitwidths at intermediate reduction stages.
A custom implementation doing this internally may yield further performance improvements; we leave this to future work.

\subsection{GPT on OpenwebText}

All our GPT training results use adaptations of the original GPT-2-style models~\citep{radford2019language}, incorporating modern training techniques from LLaMA~\citep{touvron2023llama}, such as Gated Linear Unit activations and RMSNorm.
These adjustments are aligned with the techniques in~\citet{liu2024communication}.
The learning rate was generally set to 6e-5, except for the 1558M model, which was trained with a learning rate of 1e-5. The value of $\beta_1$ was set to 0.9, and gradient clipping was applied at 1.0. Weight decay was also applied with a value of 1.0.
The batch size was typically 12, but for the 1558M model, it was reduced to 8 due to memory constraints. Gradient accumulation was not used.
The configurations for each model size are as follows:
\begin{itemize}
    \item 124M GPT: $n_{\text{layer}}=12$, $n_{\text{head}}=12$, $n_{\text{embd}}=768$
    \item 350M GPT: $n_{\text{layer}}=24$, $n_{\text{head}}=16$, $n_{\text{embd}}=1024$
    \item 774M GPT: $n_{\text{layer}}=36$, $n_{\text{head}}=20$, $n_{\text{embd}}=1280$
    \item 1558M GPT: $n_{\text{layer}}=48$, $n_{\text{head}}=25$, $n_{\text{embd}}=1600$
\end{itemize}
The sequence length was set to 1024 for all models and dropout was not applied.

\subsection{GPT on FineWeb}

We trained a 774M-parameter GPT model (36 layers, 20 attention heads, embedding dimension 1 280) for 50,000 steps . 
We used the Lion optimizer with a base learning rate of 3e-5; in addition, we scaled the head, embedding, and scalar parameter groups to 3e-3, 3e-2, and 3e-3 respectively. 
The betas were set to ($\beta_1=0.8, \beta_2=0.95$), weight decay to 1e-2, and no gradient clipping was applied. 
We used a training sequence length of 12,288 tokens and a validation sequence length of 65,536 tokens with no dropout. 

\subsection{ResNet and DeiT}

For both ResNet and DeiT models, the learning rate was generally set to 3e-4 and scheduled using a cosine scheduler. The value of $\beta_1$ was set to 0.9, with a weight decay applied at 0.0001. In addition to RandomCrop, Cutout, and RandomHorizontalFlip, we also applied auto-augmentation tailored for CIFAR.
For ResNet18, we used the CIFAR-optimized version as described in the Cutout paper \cite{devries2017improved}.



\end{document}